\documentclass[runningheads]{llncs}
\usepackage{graphicx}

\usepackage{tikz}
\usepackage{comment}
\usepackage{amsmath,amssymb} 
\usepackage{color}
\usepackage{subcaption}
\usepackage{orcidlink}

\usepackage[accsupp]{axessibility}  

\usepackage{wrapfig}
\usepackage{multirow}
\newcommand{\Method}{SPI}

\newcommand{\Loss}[1]{\mathcal{L}_{#1}}
\newcommand{\contrastiveLoss}{\mathop{\Loss{con}}}
\newcommand{\consistencyLoss}{\Loss{ida}}
\newcommand{\pawsLoss}{\Loss{ils}}
\newcommand{\classifierLoss}{\Loss{cls}}

\newcommand{\labeledSourceDataset}{\mathop{S}}
\newcommand{\unlabeledTargetDataset}{\mathop{T}}
\newcommand{\labeledTargetDataset}{\mathop{\hat{T}}}

\newcommand{\sourceDomain}{$\mathcal{D}_{s}$}
\newcommand{\targetDomain}{$\mathcal{D}_{t}$}

\newcommand{\threshold}{$\gamma$}
\newcommand{\momentum}{$\rho$}
\newcommand{\featureExtractor}{$\mathcal{F}$}
\newcommand{\featureExtractorMapping}{\featureExtractor$ : \mathcal{X} \rightarrow \mathbb{R}^{d}$}
\newcommand{\classifier}{$\mathcal{H}$}
\newcommand{\classifierMapping}{\classifier$ : \mathbb{R}^{d} \rightarrow \mathbb{R}^{C}$}

\DeclareMathOperator*{\argmax}{arg\,max}
\newcommand{\temp}{$\tau$}
\newcommand{\warmup}{$W$}
\newcommand{\nSamplesPerClass}{\mathop{\eta_{sup}}}
\newcommand{\nGlobalCrops}{\mathop{\eta_{g}}}
\newcommand{\nLocalCrops}{\mathop{\eta_{l}}}

\newcommand{\softmax}{$\sigma$}

\newcommand{\optLR}{$0.0002$}
\newcommand{\optWD}{$0.0005$}
\newcommand{\optMU}{$0.9$}

\newcommand{\gpu}{NVIDIA Tesla V100}

\begin{document}
\pagestyle{headings}
\mainmatter
\def\ECCVSubNumber{60}  

\title{Semi-Supervised Domain Adaptation by Similarity based Pseudo-label Injection} 


\titlerunning{SPI}
%

\author{
Abhay Rawat\inst{1,2}\orcidlink{0000-0003-3406-2149}\index{Rawat, Abhay} \and
Isha Dua\inst{2}\orcidlink{0000-0001-5494-059X} \and
Saurav Gupta\inst{2}\orcidlink{0000-0003-2059-1760} \and
Rahul Tallamraju\inst{2}\orcidlink{0000-0002-8087-2225}
}

%
\authorrunning{A. Rawat et al.}
%
\institute{International Institute of Information Technology, Hyderabad, India \and
Mercedes-Benz Research and Development India, Bengaluru, India\\
\email{\{firstname.lastname\}@mercedes-benz.com}
}
\maketitle

\begin{abstract}
One of the primary challenges in Semi-supervised Domain Adaptation (SSDA) is the skewed ratio between the number of labeled source and target samples, causing the model to be biased towards the source domain.
Recent works in SSDA show that aligning only the labeled target samples with the source samples potentially leads to incomplete domain alignment of the target domain to the source domain.
In our approach, to align the two domains, we leverage contrastive losses to learn a semantically meaningful and a domain agnostic feature space using the supervised samples from both domains.
To mitigate challenges caused by the skewed label ratio, we pseudo-label the unlabeled target samples by comparing their feature representation to those of the labeled samples from both the source and target domains.
Furthermore, to increase the support of the target domain, these potentially noisy pseudo-labels are gradually injected into the labeled target dataset over the course of training.
Specifically, we use a temperature scaled cosine similarity measure to assign a soft pseudo-label to the unlabeled target samples.
Additionally, we compute an exponential moving average of the soft pseudo-labels for each unlabeled sample.
These pseudo-labels are progressively injected (or removed) into the (from) the labeled target dataset based on a confidence threshold to supplement the alignment of the source and target distributions. Finally, we use a supervised contrastive loss on the labeled and pseudo-labeled datasets to align the source and target distributions.
Using our proposed approach, we showcase state-of-the-art performance on SSDA benchmarks - Office-Home, DomainNet and Office-31. The inference code is available at \href{https://github.com/abhayraw1/SPI}{https://github.com/abhayraw1/SPI}
\end{abstract}

\section{Introduction}

\begin{figure}[t]
    \centering
    \begin{subfigure}[b]{0.3\textwidth}
        \centering
        \includegraphics[scale=0.35, trim=0.cm 0cm 0.0cm 0cm, clip=true]{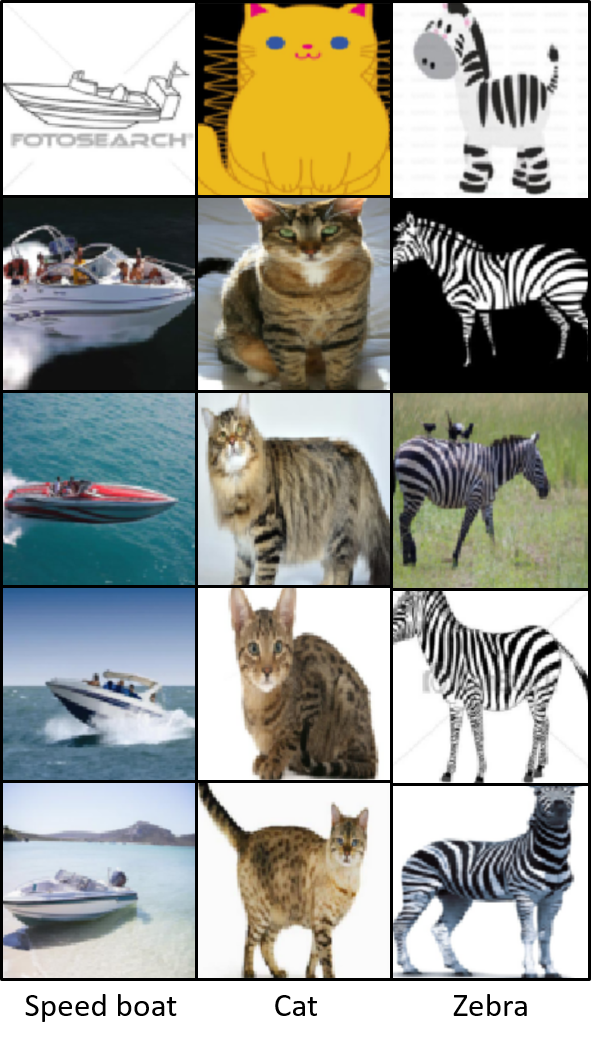}
        \caption{}
        \label{fig:nn_teaser}
    \end{subfigure}
    \hfill
    \begin{subfigure}[b]{0.68\textwidth}
        \centering
        \includegraphics[scale=0.35, trim=0.3cm 0cm 0.5cm 0cm, clip=true]{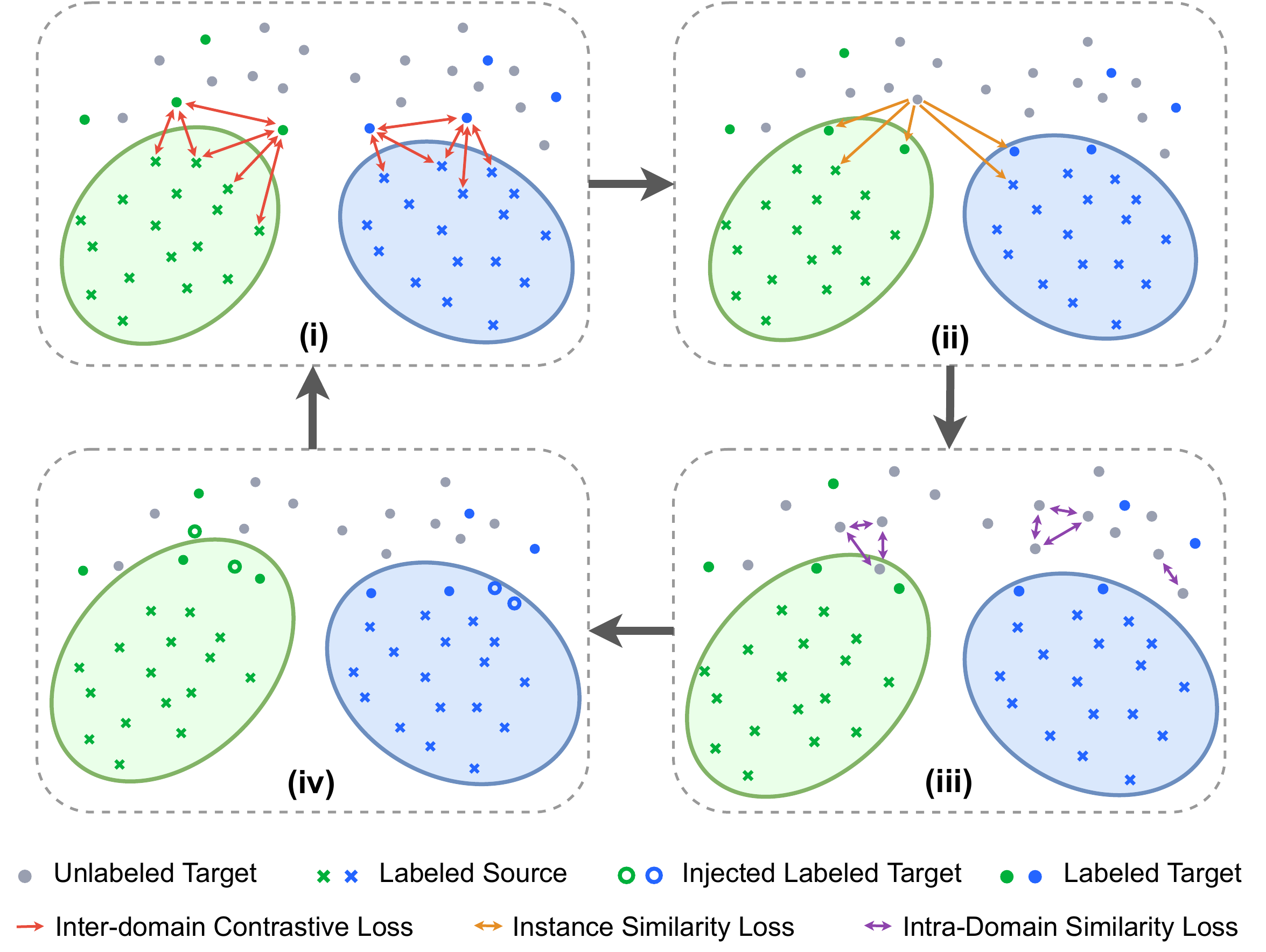}
        \caption{}
        \label{fig:teaser}
    \end{subfigure}
    \caption{(a) Nearest neighbours of the target domain (Real) from the source domain (Clipart). (b) High level overview of our proposed method \Method. (i) Inter-domain contrastive loss works on the labeled source and target samples, pulling the samples from the same class closer, (ii) unlabeled samples are drawn towards similar samples from the support set using the instance level similarity loss, (iii) Intra-domain alignment loss focuses on bringing similar samples in the unlabeled target domain closer together, (iv) Confident samples are injected into the labeled target dataset.}
\end{figure}

Domain Adaptation approaches aim at solving the data distribution shift between training (source) and test (target) data. 
Unsupervised Domain Adaptation (UDA) considers the problem of adapting the model trained on a labeled source distribution to an unlabeled target distribution.
Most recent approaches in UDA \cite{ganin2016domain,kang2019contrastive,long2018conditional,saito2017adversarial,wang2022cross} aim to learn a domain agnostic feature representation such that the features belonging to similar categories are closer together in the latent space.
The underlying assumption here is that learning a mapping to a domain agnostic feature space, along with a classifier that performs sufficiently well on the source domain, could generalize to the target domain.
However, recent studies \cite{li2021learning,wu2019domain,zhao2019learning,tachet2020domain} show that these conditions are not sufficient for successful domain adaptation and might even hurt generalization due to the discrepancy between the marginal label distributions of the two domains.

Semi-Supervised learning (SSL) \cite{assran2021paws,sohn2020fixmatch,berthelot2019mixmatch,zhang2021flexmatch} has already proven to be highly efficient in terms of performance per annotation and thus provides a more economical way to train deep learning models.
UDA approaches in general, however, do not perform well in a semi-supervised setting where we have access to some labeled samples from the target domain \cite{saito2019semi}.
Semi-Supervised Domain Adaptation (SSDA) \cite{singh2021clda,li2021cross,kim2020attract}, leverages a small amount of labeled samples in the target domain to aid in learning models with low error rate on the target domain.
However, as shown by \cite{kim2020attract}, simply aligning the labeled target samples with the labeled source samples can cause intra-domain discrepancy in the target domain.
During training, the labeled target samples get pulled towards the corresponding source sample clusters.
However, the unlabeled samples with less correlation to the labeled target samples are left behind.
This is because the number of labeled source samples dominate the number of labeled tareget samples, leading to a skewed label distribution.
This results in sub-distributions within the same class of the target domain.
To mitigate this skewed ratio between the labeled samples from the source and target domain, recent approaches \cite{kang2019contrastive,wang2022cross} assign pseudo-labels to unlabeled data.
However, these pseudo-labels are potentially noisy and may result in poor generalization to the target domain.

In this paper, we present a simple yet effective way to mitigate the above-mentioned challenges faced in SSDA. To align the \textit{supervised samples} from both the domains, we leverage a contrastive loss to learn a semantically meaningful and a domain invariant feature space.
To remedy the intra-domain discrepancy problem, we compute soft pseudo-labels for the unlabeled target samples by comparing their feature representations to those of the labeled samples.
However, samples with less correlation with the labeled ones could have noisy and incorrect pseudo-labels.
Therefore, we gradually inject (or remove) pseudo-labeled samples into (or from) the labeled target dataset throughout training, based on the model's confidence on the respective pseudo-labels.

Our proposed method, \Method \ (Similarity based Pseudo-label Injection) for SSDA has the following four components.

\begin{enumerate}
  \item \textbf{Domain alignment - between labeled source and labeled target samples:} To align the feature representations of the two domains, we use a supervised contrastive loss on the labeled samples from both domains. This enforces similar representations for samples belonging to the same class across both domains, as shown in Fig. \ref{fig:teaser} (i).
  \item \textbf{Soft Pseudo-Labeling - between different views of the unlabeled samples:} We use a non-parametric, similarity based pseudo labeling technique to generate soft pseudo-labels for the unlabeled samples using the labeled samples from both the domains as shown in Fig \ref{fig:teaser} (ii). To further enforce consistency of unlabeled target sample representations, we employ a similarity-based consistency between different augmented views of the same instance (Instance Similarity Loss).
  \item \textbf{Intra-domain similarity - between the unlabeled samples of the target domain:} We compare the positions of the most highly activated feature dimensions of the unlabeled samples to ascertain if the samples are semantically similar. We then minimize the distance between feature representations of similar samples as shown in Fig. \ref{fig:teaser} (iii).
  \item \textbf{Pseudo-label Injection:} Finally, an exponential moving average of the sample's soft pseudo-label is updated over training. Samples with pseudo-label confidence above a given threshold are injected into the labeled target dataset as shown in Fig. \ref{fig:teaser} (iv).
\end{enumerate}


In summary, we present a novel method that leverages feature level similarity between the unlabeled and labeled samples across domains to pseudo-label the unlabeled data.
We use a novel loss function that includes supervised contrastive loss, an instance-level similarity loss and an intra-domain consistency loss which together help to bring the features of similar classes closer for both domains.
Finally, we propose to gradually inject and remove pseudo-labels into the labeled target dataset to increase the support of the same.
We showcase superior classification accuracy across popular image classification benchmarks - Office-Home, Office-31, and DomainNet against state-of-the-art SSDA approaches.

\section{Related Works}
\textbf{Unsupervised Domain Adaptation (UDA)} methods \cite{ganin2016domain,chen2019joint,kang2019contrastive,long2015learning,wang2022cross,li2021cross,saito2017adversarial} aims to reduce the cross-domain divergence between the source and target domains by either minimizing the discrepancy between the two domains or, by training the domain invariant feature extractor via adversarial training regimes  \cite{chen2019joint,long2015learning,paul2020domain,ganin2015unsupervised,ganin2016domain,saito2017adversarial}. 
The distribution level distance measures have been adopted to reduce the discrepancy like Correlation Distances \cite{yao2015semi,sun2016return}, Maximum Mean Discrepancy (MMD) \cite{kang2019contrastive,long2015learning}, JS Divergence \cite{shui2020beyond} and Wasserstein Distance \cite{shen2018wasserstein}.
SSDA approaches, on the other hand, utilize a small number of labeled samples from the target domain which is a more realistic setting in most vision problems.


\textbf{Semi-Supervised Domain Adaptation}
deals with data sampled from two domains - one with labeled samples and the other with a mix of labeled and unlabeled samples - and hence aims to tackle the issues caused due to domain discrepancy.
Recent works \cite{saito2019semi,kim2020attract} show that UDA methods do not perform well when provided with some labeled samples due to the bias towards labeled samples from the source domain.
Other approaches for SSDA rely on learning domain invariant feature representations \cite{li2021cross,kim2020attract,singh2021clda,saito2019semi}.
However, studies \cite{li2021learning,wu2019domain,zhao2019learning,tachet2020domain} have shown that learning such a domain agnostic feature space and a predictor with low error rate is not a sufficient condition to assure generalization to the target domain.


\textbf{Pseudo Labeling} techniques have been used in UDA and SSDA approaches to reduce the imbalance in the labeled source and target data.
CAN \cite{kang2019contrastive} and CDCL \cite{wang2022cross} uses $k$-means approach to assign pseudo labels to the unlabeled target samples.
D{\footnotesize E}C{\footnotesize O}T{\footnotesize A} \cite{yang2021deep} and CLDA \cite{singh2021clda} use the predictions of their classifiers as the pseudo-label.
CDAC \cite{wang2022cross} uses confident predictions of the weakly augmented unlabeled samples as pseudo-labels for the strongly augmented views of the same sample, similar to FixMatch \cite{sohn2020fixmatch}.
In contrast, our approach uses a similarity based pseudo labeling approach inspired by PAWS \cite{assran2021paws}.
We extend this approach to the SSDA setting by computing the pseudo-labels for the unlabeled samples using labeled samples from both the domains.

\textbf{Contrastive Learning} aims to learn feature representations by pulling the positive samples closer and pushing the negative samples apart \cite{bachman2019learning,van2018representation,tian2020contrastive,long2015learning,he2020momentum}.
In a supervised setting, \cite{khosla2020supervised} leverages the labels of samples in a batch to construct positive and negative pairs.
But, due to the absence of labeled samples in UDA, methods like CDCL \cite{wang2022cross}, assign pseudo labels to the unlabeled target samples using clustering to help construct the positive and negative pairs.
CLDA \cite{singh2021clda}, an SSDA method, uses contrastive loss to align the strongly and weakly augmented views of the same unlabeled image to ensure feature level consistency.

\section{Method}


In semi-supervised domain adaptation (SSDA), apart from labeled samples from the source distribution \sourceDomain \ and the unlabeled samples from the target distribution \targetDomain, we also have access to a small amount of labeled samples from \targetDomain.
Let $ \labeledSourceDataset = \{ (x_{i}^{s},  y_{i}^{s}) \}_{i=1}^{N_{s}}$ and $ \unlabeledTargetDataset = \{ (x_{i}^{t}) \}_{i=1}^{N_{u}}$ denote the set of data sampled i.i.d. from \sourceDomain \ and \targetDomain \ respectively.
Additionaly, we also have $ \labeledTargetDataset = \{ (x_{i}^{t}, y_{i}^{t}) \}_{i=1}^{N_{l}}$, a small set of labeled samples from \targetDomain.
The number of samples in the $\labeledSourceDataset$, $\unlabeledTargetDataset$ \ and $\labeledTargetDataset$ \ are denoted by $N_{s}$, $N_{u}$, $N_{l}$ respectively, where $N_{l} \ll N_{u}$
Both domains are considered to have the same set of labels $Y = \{1, 2, \cdots C\}$ where $C$ is the number of categories.
Our aim is to learn a classifier using $\labeledSourceDataset$, $\unlabeledTargetDataset$ \ and $\labeledTargetDataset$ \ that can accurately classify novel samples from the target domain \targetDomain \ during inference/testing.

\subsubsection*{\textbf{Preliminaries:}} Our method consists of a feature extractor \featureExtractorMapping \ and a classifier \classifierMapping \ parameterized by $\Theta_{\mathcal{F}}$ and $\Theta_{\mathcal{H}}$ respectively, where $\mathcal{X}$ denotes the image space, $d$ denotes the dimensionality of the feature space and $C$ is the number of classes.
Classifier \classifier \ is a single linear layer that maps the features from the feature extractor \featureExtractor \ to unnormalized class scores.
These scores are converted into respective class probabilities using the softmax function \softmax$(\cdot)$.

\begin{equation}
  \label{eqn:cls_prob}
  h(x) = \sigma(\mathcal{H}(z)),
\end{equation}
where $z = \mathcal{F}(x)$ denotes the feature representation of the input image $x$.
For brevity, we denote the class probability distribution for a sample $x_{i}$ computed using the classifier as ${h}_{i}$.

\begin{figure*}[t]
  \includegraphics[width=\textwidth,trim={0 3cm 0 0},clip]{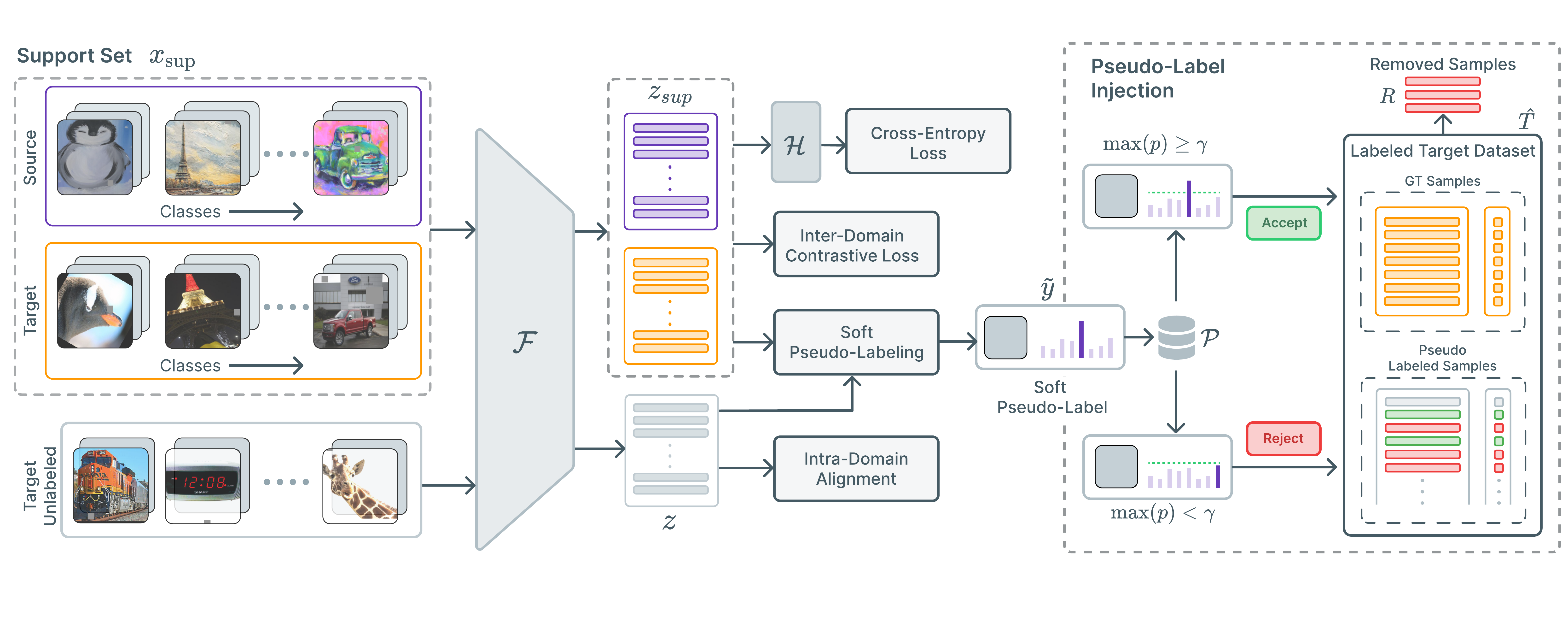}
  \caption{An overview of our proposed method, \Method.
    We sample labeled images from both the source (orange) and target (blue) domain to construct the support set $x_{sup}$.
    Features from the unlabeled samples $z$ are compared with those from the support set $z_{sup}$ to compute the soft pseudo-labels $\tilde{y}$ for the target unlabeled images.
    $\mathcal{P}$ maintains a mapping from the image ids to the exponential moving averages of these soft pseudo-labels.
    After every epoch, we select samples from $\mathcal{P}$ for injection (green) into the target labeled dataset $\hat{T}$ based on the confidence threshold $\gamma$ (if $p = \mathcal{P}(\operatorname{ID}(.)) \geq \gamma$).
    Additionally, injected samples are also selected for removal (red) from $\hat{T}$ if their confidence falls below $\gamma$. Classifier is denoted as $\mathcal{H}$
  }
  \label{fig:block_diagram}
\end{figure*}

Fig. \ref{fig:block_diagram} depicts a concise overview of our proposed method, \Method.
We sample $\nSamplesPerClass$ images per class (with replacement) from the labeled source $\labeledSourceDataset$ and the target dataset $\labeledTargetDataset$ to construct the mini-batch of labeled samples for each domain.
We refer to the concatenated mini-batch of labeled samples from both domains as the \textit{support set} throughout the paper.
The support set contains $\nSamplesPerClass C$ samples from both the domains, $2 \nSamplesPerClass C$ samples in total.
For the unlabeled batch $B_{u}$, we sample images uniformly from the unlabeled target dataset $\unlabeledTargetDataset$.
Following the multi-view augmentation technique \cite{assran2021paws,caron2020unsupervised,caron2021emerging}, each image in the unlabeled batch $B_{u}$ is further augmented to produce $\nGlobalCrops$ \ global views and $\nLocalCrops$ \ local views.
Throughout the paper, we use the term \textit{unlabeled images} to refer to the global views of these unlabeled images, unless specified otherwise.
For the sake of simplicity, we use a common symbol $\tau$ to denote the temperature in different equations, Eqn. \ref{eqn:contrastive}, \ref{eqn:pseudo_label} and \ref{eqn:sharpening}.

\subsubsection*{\textbf{An outline of \Method}}
In Sec. \ref{sec:inter_domain_alignment}, we introduce the supervised contrastive loss used for domain alignment.
We explain the process of injecting unlabeled samples into the labeled dataset in detail in Sec. \ref{sec:pseudo_labeling}.
Sec. \ref{sec:instance_level_consistency} presents our modification of the multi-view similarity loss \cite{assran2021paws}, used in SSL, to the SSDA setting.
This loss helps in two ways: a) by ensuring the feature representation for different views of the same image is similar, and b) by bringing these representations closer to similar samples from the support set.
In Sec. \ref{sec:intra_domain_similarity}, we present a technique to determine if two samples from the unlabeled target domain are similar at a feature level \cite{han2020automatically}.
Subsequently, we present a feature-level similarity loss between similar samples to bring them closer to each other in the latent space.
Finally, we combine these components, presenting \Method \ as an end-to-end framework in Sec. \ref{sec:framework}.

\subsection{Inter-Domain Feature Alignment}
\label{sec:inter_domain_alignment}
The premise for most domain adaptation approaches is to learn a domain invariant feature representation for both the source and target domains.
CDCL \cite{wang2022cross} uses contrastive learning in the context of UDA to align the source and target domains.
Due to the absence of class labels in UDA, they use pseudo-labels generated using k-means clustering for the contrastive loss.
CLDA \cite{singh2021clda}, a method for SSDA, applies contrastive loss on the exponential moving average of the source and target feature centroids to align the two domains.
In contrast, we leverage contrastive loss more directly by explicitly treating the samples of the same classes as positives, irrespective of the domain.
The feature extractor is then trained to minimize the $\contrastiveLoss$ \ by maxmizing the similarity between the features of the same class.


Let $A = \{1, 2, \cdots, 2\nSamplesPerClass C\}$ denote the set of indices in the support set.
Note that the support set contains samples from both the domains and $\nSamplesPerClass$ samples from each class.
Let $P_{i}$ denote the set indices of all the images with the same label as $i$.
Then, the supervised contrastive loss $\contrastiveLoss$ \ can be written as:

\begin{equation}
  \label{eqn:contrastive}
  \contrastiveLoss = \sum_{i \in A} \frac{-1}{|P_{i}|} \sum_{p \in P_{i}} \log \frac{\exp(z_{i} \cdot z_{p}/\tau)}{\sum\limits_{a \in A \setminus i} \exp(z_{a} \cdot z_{p}/\tau)} \ ,
\end{equation}
where, \temp \ denotes the temperature and controls the compactness of the clusters for each class, and $z = \mathcal{F}(\cdot)$ is the feature embeddings of the input images.

However, as pointed out in \cite{kim2020attract}, aligning the labeled samples from the source and target domains can lead to sub-distributions in the target domain.
More specifically, the unlabeled samples with less correlation to the labeled samples in the target domain do not get pulled by the contrastive loss.
This causes intra-domain discrepancy and thereby leads to poor performance.
To mitigate this issue, we propose to inject unlabeled samples into the labeled target dataset $\labeledTargetDataset$ \ thus, effectively increasing the support of the labeled samples in the target domain.
We discuss this approach in more detail Sec. \ref{sec:pseudo_labeling}.

\subsection{Pseudo-Label Injection}
\label{sec:pseudo_labeling}
To reduce the intra-domain discrepancy, we propose injecting samples from the unlabeled target dataset $\unlabeledTargetDataset$ \ into the labeled target dataset $\labeledTargetDataset$.
Using the support set, we first compute the soft pseudo-labels of the unlabeled samples.
Throughout the training, we keep an exponential moving average of the sharpened soft pseudo-labels for every sample in the unlabeled target dataset $\unlabeledTargetDataset$.
This moving average estimates the confidence of our model's prediction of each unlabeled sample.
Using this estimate, we inject the highly confident samples into the labeled target dataset $\labeledTargetDataset$ with their respective label set to the dominant class after each epoch.

To compute the soft pseudo-labels for the unlabeled samples from the target domain, we take inspiration from PAWS \cite{assran2021paws}, a recent work in semi-supervised learning and extend it to the SSDA setting.
We denote the support set $x_{sup}$ and their respective labels as $y_{sup}$.
Let $\hat{z}_{sup}$ be the normalized features representation of the samples in the support set $x_{sup}$ and $\hat{z}_{i} ( = {z_{i}} / {\lVert z_{i} \rVert} )$ denote the normalized feature representation for the unlabeled sample $x_{i}$.
Then, the soft pseudo-label for $i$th unlabeled sample can be computed using:
\begin{equation}
  \label{eqn:pseudo_label}
  \tilde{y}_{i} = \sigma_{\tau}(\hat{z}_{i} \cdot \hat{z}_{sup}^{\top}) \ y_{sup}
\end{equation}
where $\sigma_{\tau}(\cdot)$ denotes the softmax operator with temperature $\tau$.
These soft pseudo-labels are then sharpened using the sharpening function $\pi$ with temperature $\tau > 0$, described as follows:
\begin{equation}
  \label{eqn:sharpening}
  \pi(\tilde{y}) = \frac{\tilde{y}^{1 / \tau}}{\sum_{j=1}^{C} \tilde{y}_{j}^{1 / \tau}}
\end{equation}
Sharpening helps to produce confident predictions from the similarity measure between the unlabeled and labeled samples.

Throughout the training, we keep an exponential moving average (EMA) of the sharpened soft pseudo-labels of each image in the unlabeled target dataset $\unlabeledTargetDataset$.
More specifically, we maintain a mapping $\mathcal{P}: \mathbb{I} \rightarrow \mathbb{R}^C$ from image ids of the unlabeled samples to the running EMA of their respective sharpened soft pseudo-labels (class probability distribution).
Let $\operatorname{ID}(\cdot)$ denote an operator that returns the image-id corresponding to the input sample in the unlabeled target dataset $\unlabeledTargetDataset$ and $\mathcal{P}(\operatorname{ID}(x_{i}))$ be the EMA of the sharpened pseudo-label of $x_{i}$.
Then, this exponential moving average for a sample $x_{i}$ in the unlabeled dataset $\unlabeledTargetDataset$ is updated as follows:
\begin{equation}
  \label{eqn:ema}
  \mathcal{P}(\operatorname{ID}(x_{i})) \leftarrow \rho \ \pi(\tilde{y}_{i}) + (1 - \rho) \ \mathcal{P}(\operatorname{ID}(x_{i}))
\end{equation}
where $\rho$ denotes the momentum parameter.
When a sample is encountered for the first time in the course of training, $\mathcal{P}(\operatorname{ID}(x_{i}))$ is set to $\pi(\tilde{y}_{i})$ and Eqn. \ref{eqn:ema} is used thereafter.

After each epoch, we examine the EMA (class probability distribution) for each sample in $\mathcal{P}$.
If the confidence of a particular sample for a class crosses a certain threshold \threshold, we inject that sample and its corresponding predicted class into the labeled target dataset $\labeledTargetDataset$.
We define the set of samples considered for injection $I$ as:
\begin{equation}
  \label{eqn:set_inject}
  I_{t} \triangleq \{ (x_{i}, \argmax \mathcal{P}(\operatorname{ID}(x_{i})) \ \vert \ x_{i} \in  T \land \max{\mathcal{P}(\operatorname{ID}(x_{i}))} \geq \gamma \} \ ,
\end{equation}
where $t$ denotes the current epoch.

However, these samples could potentially be noisy and might hinder the training process; therefore we also remove samples from the labeled dataset $\labeledTargetDataset$ if their confidence falls below the threshold \threshold.
The set of samples to be removed from the labeled target dataset $R$ is defined as:

\begin{equation}
  \label{eqn:set_reject}
  R_{t} \triangleq \{ (x_{i}, y_{i}) \ \vert \ x_{i} \in  (\hat{T}_{t} \setminus \hat{T}_{0}) \land \max{\mathcal{P}(\operatorname{ID}(x_{i}))} < \gamma \} \ ,
\end{equation}

where $y_{i}$, denotes the corresponding pseudo-label that had been previously assigned to the sample $x_{i}$ from Eqn. \ref{eqn:set_inject}.
Note that the original samples from the labeled target dataset $\labeledTargetDataset_{0}$ are never removed from the dataset as both $I$ and $R$ contain samples only from the unlabeled target dataset $\unlabeledTargetDataset$.

The labeled target dataset $\labeledTargetDataset$ after each epoch $t$, is therfore updated as:
\begin{equation}
  \label{eqn:label_update}
  \hat{T}_{t+1} =
  \begin{cases}
    (\hat{T}_{t} \setminus R_{t} ) \cup I_{t} & \text{if} \ t \geq W \\
    \hat{T}_{t}                               & \text{otherwise}
  \end{cases}
\end{equation}
where $W$ represents the number of warmup epochs up to which the labeled target dataset $\labeledTargetDataset$ \ remains unaltered.
These warmup epochs allow the feature representations of the source and target domains to be aligned to some extent before the samples are injected into the label target dataset.
This prevents false-positive samples from getting into $\labeledTargetDataset$ which would otherwise hinder the learning process.

\subsection{Instance Level Similarity}
\label{sec:instance_level_consistency}
We now introduce the Instance Level Similarity loss.
Inspired by \cite{assran2021paws,caron2021emerging}, we follow a multi-view augmentation to generate $\nGlobalCrops=2$ global crops and $\nLocalCrops$ local crops of the unlabeled images.
The key insight behind such an augmentation scheme is to enforce the model to focus on the object of interest by explicitly bringing the feature representations of these different views closer.
The global crops contain more semantic information about the object of interest, whereas the local crops only contain a limited view of the image (or object).
By computing the feature level similarity between the global crops and the support set samples, we compute the pseudo-label for the unlabeled samples using Eqn. \ref{eqn:pseudo_label}.

The feature extractor is then trained to minimize the cross-entropy between the pseudo-label generated using one global view and the sharpened pseudo-label generated using the other global view.
Additionally, the cross-entropy between the pseudo-label generated using the local views and the mean of the sharpened pseudo-label from the global views is added to the loss.

With a slight abuse of notation, given a sample $x_{i}$, we define $\tilde{y}_{i}^{g_1}$ and $\tilde{y}_{i}^{g_2}$ as the pseudo-label for the two global crops and $\tilde{y}_{i}^{l_{j}}$ denote the pseudo-label for the $j$th local crop.
Similarly, we follow the same notation to define the sharpened pseudo-label for these crops denoted by $\pi$.
The feature extractor is thus trained to minimize the following loss:
\begin{equation}
  \label{eqn:paws_loss}  \pawsLoss = - \sum_{i=1}^{\vert B_{u} \vert} \Big( \operatorname{H}(\tilde{y}_{i}^{g_1}, \pi_{i}^{g_2}) + \operatorname{H}(\tilde{y}_{i}^{g_2}, \pi_{i}^{g_1}) + \sum_{j=1}^{\nLocalCrops} \operatorname{H}(\tilde{y}_{i}^{l_{j}}, \pi_{i}^{g}) \Big) \ ,
\end{equation}
where, $\operatorname{H}(\cdot, \cdot)$ denotes the cross-entropy, $\pi_{i}^{g} = (\pi_{i}^{g_1}  + \pi_{i}^{g_1}) / 2 $ and $\vert B_{u} \vert$ denotes the number of unlabeled samples.

\subsection{Intra-Domain Alignment}
\label{sec:intra_domain_similarity}
To ensure that the unlabeled samples from the same class in the target domain are closer together in the latent space, we use consistency loss between the unlabeled samples.
Due to the absence of labels for these samples, we compute the pairwise feature similarity between the unlabeled samples to estimate whether they could potentially belong to the same class.
As proposed by \cite{han2020automatically}, two samples $x_{i}$ and $x_{j}$ can be considered similar if the indices of their top-k highly activated feature dimensions are the same.
Let $\operatorname{top-k} \ (z)$ denote the set of indices of the top $k$ highly activated feature dimensions of $z$ then, we consider two unlabeled samples $i$ and $j$ similar if:
\begin{equation}
  \label{eqn:unlb_similarity}
  \operatorname{top-k} \ (z_{i}) \ominus \operatorname{top-k} \ (z_{j}) = \Phi
\end{equation}
where, $z_{i}$ and $z_{j}$ are their respective feature representations, and $\ominus$ is the symmetric set difference operator.

We construct a binary matrix $M \in \{0, 1\}^{\vert B_{u} \vert \times \vert B_{u} \vert}$ whose individual entries $M_{ij}$ denote whether the $i$th sample is similar to $j$th sample in the unlabeled batch $B_{u}$ using Eqn. \ref{eqn:unlb_similarity}.
Using the similarity matrix $M$, we compute the intra-domain consistency loss $\consistencyLoss$ \ for the target unlabeled samples as follows:
\begin{equation}
  \consistencyLoss = \frac{1}{\vert B_{u} \vert^{2}} \sum_{i=1}^{\vert B_{u} \vert} \sum_{j=1}^{\vert B_{u} \vert} M_{ij} \ \lVert z_{i} - z_{j} \rVert_{2}
\end{equation}

\subsection{Classification Loss and Overall Framework}
\label{sec:framework}
We use the label-smoothing cross-entropy \cite{muller2019does} loss to train the classifier layer.
For the classifier training, we only use the samples from the labeled source dataset $\labeledSourceDataset$ \ and the labeled target dataset $\labeledTargetDataset$, which is constantly being updated with new samples.
\begin{equation}
  \label{eqn:xent}
  \classifierLoss = - \sum_{i=1}^{2 \nSamplesPerClass C} \operatorname{H}(h_{i}, \hat{y}_{i})
\end{equation}
where, $h_{i}$ is the predicted class probabilities (Eqn. \ref{eqn:cls_prob}), $\operatorname{H}$ denotes the cross-entropy loss and $\hat{y}_{i} = (1 - \alpha) y_{i} + \alpha/C$ is the smoothened label corresponding to $x_{i}$.
Here, $\alpha$ is the smoothing parameter and $y_{i}$ is the one-hot encoded label vector.

Combining the different losses used in our proposed method \Method, $\contrastiveLoss$, $\pawsLoss$ \ and $\consistencyLoss$, yields a single training objective:
\begin{equation}
  \label{eqn:final_loss}
  \mathcal{L}_{\Method} = \lambda \contrastiveLoss + \pawsLoss + \consistencyLoss + \classifierLoss
\end{equation}

\section{Experiments}

\subsection{Datasets}
We compare the performance of \Method \ against the existing approaches on the popular image classification benchmarks: Office-Home \cite{venkateswara2017deep}, Office-31 \cite{saenko2010adapting} and DomainNet \cite{peng2019moment}.
Office-Home comprises of $4$ domains: \textbf{A}rt, \textbf{C}lipart, \textbf{P}roduct, and \textbf{R}ealWorld, and a total of $65$ different categories.
The Office-31 benchmark contains $3$ domains - \textbf{A}mazon, \textbf{W}ebcam, and \textbf{D}SLR, containing objects from $31$ different categories.
Following \cite{kim2020attract,singh2021clda}, we use a subset of the DomainNet benchmark adapted by \cite{saito2019semi}, which contains a total of $126$ classes and $4$ domains: \textbf{C}lipart, \textbf{S}ketch, \textbf{P}ainting, and \textbf{R}eal.

Following the authors of \cite{kim2020attract,singh2021clda,li2021cross}, we use the publically available train, validation, and test splits provided by \cite{saito2019semi} to ensure a consistent and fair comparison against the existing approaches.


\subsection{Baselines}
To quantify the efficacy of our approach, we compare its performance to the previous state-of-the-art methods in SSDA task: \textbf{CLDA} \cite{singh2021clda}, \textbf{CDAC} \cite{li2021cross}, \textbf{MME} \cite{saito2019semi}, \textbf{APE} \cite{kim2020attract}, \textbf{D{\footnotesize E}C{\footnotesize O}T{\footnotesize A}} \cite{yang2021deep}, \textbf{BiAT} \cite{jiang2020bidirectional}, \textbf{UODA} \cite{qin2021contradictory}, \textbf{ENT} \cite{grandvalet2004semi} and \textbf{Meta-MME} \cite{qiu2021meta}.
We also include the results of popular UDA approaches \textbf{DANN} \cite{ganin2016domain}, \textbf{ADR} \cite{saito2017adversarial} and \textbf{CDAN} \cite{long2018conditional}.
These UDA methods have been modified to utilize the labeled target samples to ensure a fair comparison, as reported in \cite{saito2019semi}.
In addition to this, \textbf{S+T} method provides a rudimentary baseline wherein the model is trained only using the labeled samples from both the source and target domains.

\begin{table*}[t]
    \centering
    \caption{Accuracy on Office-Home in the 3-shot setting (ResNet-34)}
    \resizebox{\textwidth}{!}{%
        \begin{tabular}{c|cccccccccccc|c}
            \hline
            Method                                                                     & R $\rightarrow$ C & R $\rightarrow$ P & \multicolumn{1}{l}{R $\rightarrow$ A} & P $\rightarrow$ R & \multicolumn{1}{l}{P $\rightarrow$ C} & P $\rightarrow $A & A $\rightarrow$ P & \multicolumn{1}{l}{A $\rightarrow$ C} & \multicolumn{1}{l}{A $\rightarrow$ R} & \multicolumn{1}{l}{C $\rightarrow$ R} & \multicolumn{1}{l}{C $\rightarrow$ A} & C $\rightarrow$ P & Mean           \\ \hline
            S + T                                                                      & 55.7              & 80.8              & 67.8                                  & 73.1              & 53.8                                  & 63.5              & 73.1              & 54.0                                  & 74.2                                  & 68.3                                  & 57.6                                  & 72.3              & 66.2           \\
            DANN    \cite{ganin2016domain}                                             & 57.3              & 75.5              & 65.2                                  & 51.8              & 51.8                                  & 56.6              & 68.3              & 54.7                                  & 73.8                                  & 67.1                                  & 55.1                                  & 67.5              & 63.5           \\
            ENT   \cite{grandvalet2004semi}                                            & 62.6              & 85.7              & 70.2                                  & 79.9              & 60.5                                  & 63.9              & 79.5              & 61.3                                  & 79.1                                  & 76.4                                  & 64.7                                  & 79.1              & 71.9           \\
            MME   \cite{saito2019semi}                                                 & 64.6              & 85.5              & 71.3                                  & 80.1              & 64.6                                  & 65.5              & 79.0              & 63.6                                  & 79.7                                  & 76.6                                  & 67.2                                  & 79.3              & 73.1           \\
            Meta-MME   \cite{qiu2021meta}                                              & 65.2              & -                 & -                                     & -                 & 64.5                                  & 66.7              & -                 & 63.3                                  & -                                     & -                                     & 67.5                                  & -                 & -              \\
            APE \cite{kim2020attract}                                                  & 66.4              & 86.2              & 73.4                                  & 82.0              & 65.2                                  & 66.1              & 81.1              & 63.9                                  & 80.2                                  & 76.8                                  & 66.6                                  & 79.9              & 74.0           \\
            D{\footnotesize E}C{\footnotesize O}T{\footnotesize A} \cite{yang2021deep} & \textbf{70.4}     & \textbf{87.7}     & 74.0                                  & 82.1              & \textbf{68.0}                         & 69.9              & 81.8              & 64.0                                  & 80.5                                  & 79.0                                  & 68.0                                  & 83.2              & 75.7           \\
            CDAC \cite{li2021cross}                                                    & 67.8              & 85.6              & 72.2                                  & 81.9              & \textbf{67.0}                         & 67.5              & 80.3              & 65.9                                  & 80.6                                  & 80.2                                  & 67.4                                  & 81.4              & 74.2           \\
            CLDA \cite{singh2021clda}                                                  & 66.0              & 87.6              & \textbf{76.7}                         & \textbf{82.2}     & 63.9                                  & \textbf{72.4}     & 81.4              & 63.4                                  & 81.3                                  & 80.3                                  & 70.5                                  & 80.9              & 75.5           \\
            \textbf{\Method \ (Ours)}                                                  & 69.3              & 86.9              & 74.3                                  & 81.9              & 66.6                                  & 68.6              & \textbf{82.5}     & \textbf{66.4}                         & \textbf{81.6}                         & \textbf{80.9}                         & \textbf{71.1}                         & \textbf{83.8}     & \textbf{76.15} \\ \hline
        \end{tabular}
    }
    \label{Tab:office_home}
\end{table*}
\subsection{Implementation Details}
We use ResNet-34 \cite{he2016deep} as the feature extractor in our experiments on Office-Home and DomainNet datasets and VGG-16 for experiments on the Office-31 dataset.
The feature extractors use pre-trained ImageNet \cite{deng2009imagenet} weights as provided by the PyTorch Image Models library \cite{rw2019timm}.
The last layer of the feature extractor is replaced with a linear classification layer according to the dataset of interest.

We set $\lambda = 4.0$ in Eqn. \ref{eqn:final_loss} to prioritize the effect of cross-domain contrastive loss.
The momentum parameter \momentum \ for the EMA update in Eqn. \ref{eqn:ema} is set to $0.7$ for all the experiments.
The value of the injection threshold \threshold \ in Eqn. \ref{eqn:set_inject} and Eqn. \ref{eqn:set_reject} is set to $0.8$ for the experiments on the Office-Home dataset.
For experiments on DomainNet and Office-31, we use a value of $0.9$ for the threshold.
The label smoothing parameter $\alpha$ is set to $0.1$ for all the experiments.


In our experiments, we set the number of warmup epochs \warmup $=5$ \  in Eqn. \ref{eqn:label_update} for the experiments.
For Office-Home and Office-31, we set samples-per-class ($\nSamplesPerClass$) $= 4$ and for DomainNet $\nSamplesPerClass = 2$.
We use $2$ global $\nGlobalCrops$ and $4$ local $\nLocalCrops$ crops for all the datasets.
We set the batch size of unlabeled samples $B_{u}$ as $128$ for Office-Home and DomainNet, and $32$ for Office-31.

Similar to \cite{saito2019semi}, we use a Stochastic Gradient Descent optimizer with a learning rate of \optLR, a weight decay of \optWD \, and a momentum of \optMU.
The learning rate is increased linearly from $\approx 0$ to its maximum value of \optLR \ during the \warmup \ warmup epochs.
Subsequently, it is decayed to a minimum value of $10^{-5}$ using a cosine scheduler during training.
We utilize RandAugment \cite{cubuk2020randaugment} as the augmentation module in our setup.
Specifically, we use the implementation of RandAugment provided by \cite{li2021cross} for our experiments and comparisons.

All experiments for Office-Home and Office-31 were done on a single \gpu \ GPU, whereas we used a distributed setup with $2$ GPUs for DomainNet experiments.
For more experiments and more detailed report on the implementation details, we refer the reader to the supplementary report.

\begin{table*}[t]
    \centering
    \caption{Accuracy on DomainNet in 1-shot and 3-shot settings (ResNet-34)}
    \label{Tab:domainnet}
    \resizebox{\textwidth}{!}{%
        \begin{tabular}{c|cccccccccccccc|cc}
            \hline
            \multirow{2}{*}{Method}   & \multicolumn{2}{c}{R→C}    & \multicolumn{2}{c}{R→P}    & \multicolumn{2}{c}{P→C}    & \multicolumn{2}{c}{C→S}    & \multicolumn{2}{c}{S→P}    & \multicolumn{2}{c}{R→S}    & \multicolumn{2}{c|}{P→R}   & \multicolumn{2}{c}{Mean}                                                                                                                                                                                                                                            \\
                                      & \multicolumn{1}{l}{1-shot} & \multicolumn{1}{l}{3-shot} & \multicolumn{1}{l}{1-shot} & \multicolumn{1}{l}{3-shot} & \multicolumn{1}{l}{1-shot} & \multicolumn{1}{l}{3-shot} & \multicolumn{1}{l}{1-shot} & \multicolumn{1}{l}{3-shot} & \multicolumn{1}{l}{1-shot} & \multicolumn{1}{l}{3-shot} & \multicolumn{1}{l}{1-shot} & \multicolumn{1}{l}{3-shot} & \multicolumn{1}{l}{1-shot} & \multicolumn{1}{l|}{3-shot} & \multicolumn{1}{l}{1-shot} & \multicolumn{1}{l}{3-shot} \\ \hline
            S+T                       & 55.6                       & 60                         & 60.6                       & 62.2                       & 56.8                       & 59.4                       & 50.8                       & 55                         & 56                         & 59.5                       & 46.3                       & 50.1                       & 71.8                       & 73.9                        & 56.9                       & 60                         \\
            DANN     \cite{ganin2016domain}                 & 58.2                       & 59.8                       & 61.4                       & 62.8                       & 56.3                       & 59.6                       & 52.8                       & 55.4                       & 57.4                       & 59.9                       & 52.2                       & 54.9                       & 70.3                       & 72.2                        & 58.4                       & 60.7                       \\
            ADR           \cite{saito2017adversarial}              & 57.1                       & 60.7                       & 61.3                       & 61.9                       & 57                         & 60.7                       & 51                         & 54.4                       & 56                         & 59.9                       & 49                         & 51.1                       & 72                         & 74.2                        & 57.6                       & 60.4                       \\
            CDAN        \cite{long2018conditional}               & 65                         & 69                         & 64.9                       & 67.3                       & 63.7                       & 68.4                       & 53.1                       & 57.8                       & 63.4                       & 65.3                       & 54.5                       & 59                         & 73.2                       & 78.5                        & 62.5                       & 66.5                       \\
            ENT        \cite{grandvalet2004semi}                & 65.2                       & 71                         & 65.9                       & 69.2                       & 65.4                       & 71.1                       & 54.6                       & 60                         & 59.7                       & 62.1                       & 52.1                       & 61.1                       & 75                         & 78.6                        & 62.6                       & 67.6                       \\
            MME        \cite{saito2019semi}               & 70                         & 72.2                       & 67.7                       & 69.7                       & 69                         & 71.7                       & 56.3                       & 61.8                       & 64.8                       & 66.8                       & 61                         & 61.9                       & 76.1                       & 78.5                        & 66.4                       & 68.9                       \\
            UODA         \cite{qin2021contradictory}              & 72.7                       & 75.4                       & 70.3                       & 71.5                       & 69.8                       & 73.2                       & 60.5                       & 64.1                       & 66.4                       & 69.4                       & 62.7                       & 64.2                       & 77.3                       & 80.8                        & 68.5                       & 71.2                       \\
            Meta-MME    \cite{qiu2021meta}              & \textbf{-}                 & 73.5                       & \textbf{-}                 & 70.3                       & -                          & 72.8                       & \textbf{-}                 & 62.8                       & \textbf{-}                 & 68                         & \textbf{-}                 & 63.8                       & \textbf{-}                 & 79.2                        & -                          & 70.1                       \\
            BiAT        \cite{jiang2020bidirectional}              & 73                         & 74.9                       & 68                         & 68.8                       & 71.6                       & 74.6                       & 57.9                       & 61.5                       & 63.9                       & 67.5                       & 58.5                       & 62.1                       & 77                         & 78.6                        & 67.1                       & 69.7                       \\
            APE  \cite{kim2020attract}                  & 70.4                       & 76.6                       & 70.8                       & 72.1                       & 72.9                       & 76.7                       & 56.7                       & 63.1                       & 64.5                       & 66.1                       & 63                         & 67.8                       & 76.6                       & 79.4                        & 67.6                       & 71.7                       \\
            D{\footnotesize E}C{\footnotesize O}T{\footnotesize A}  \cite{yang2021deep}                  & \textbf{79.1}                       & \textbf{80.4}                       & 74.9                       & 75.2                       & 76.9                       & 78.7                       & 65.1                       & 68.6                       & 72.0                       & 72.7                       & 69.7                        & 71.9                       & 79.6                       & 81.5                        & 73.9                       & 75.6                       \\
            CDAC     \cite{li2021cross}          &       77.4         &     79.6              & 74.2                       & 75.1                       & 75.5                       & \textbf{79.3}              & 67.6                       & 69.9                       & 71                         & 73.4                       & 69.2                       & 72.5                       & 80.4                       & 81.9                        & 73.6                       & 76                         \\
            CLDA    \cite{singh2021clda}                  & 76.1                       & 77.7                       & 75.1                       & 75.7                       & 71                         & 76.4                       & 63.7                       & 69.7                       & 70.2                       & 73.7                       & 67.1                       & 71.1                       & 80.1                       & 82.9                        & 71.9                       & 75.3                       \\
            \textbf{\Method \ (Ours)} & 76.56                      & 79.2                       & \textbf{75.6}              & \textbf{76.16}             & \textbf{77.13}             & 79.2                       & \textbf{72.25}             & \textbf{72.81}             & \textbf{72.94}             & \textbf{74.5}              & \textbf{73.0}              & \textbf{73.5}              & \textbf{81.8}              & \textbf{83.2}               & \textbf{75.61}             & \textbf{76.94}             \\ \hline
        \end{tabular}%
    }
\end{table*}
\subsection{Results}
\subsubsection*{\textbf{Office-Home:}} Table \ref{Tab:office_home} compares the results of the baseline methods with our approach on the Office-Home benchmark.
We use a $3$-shot setting, as commonly used in most baseline approaches.
We observe state-of-the-art performance across the majority of the domain adaptation scenarios.
On average, we outperform the existing approaches in terms of classification accuracy.
Despite using a $3$-shot setting ($195$ images), we observe an improved performance of $5.87\%$ on A $\rightarrow$ R, $3.45\%$ on R $\rightarrow$ P, and $3.76\%$ on P $\rightarrow$ C over LIRR \cite{li2021learning}, which uses $5\%$ ($> 210$ images) of labeled target data.

\begin{wraptable}[11]{r}{5.25cm}
    \vspace{-2.5em}
        \caption{Accuracy on Office-31 under the 3-shot setting using VGG-16 backbone.}
        \label{Tab:office31}
        \begin{tabular}{c|cc|c}
            \hline
            Method                              & W→A           & D→A           & Mean          \\ \hline
            S+T                                 & 73.2          & 73.3          & 73.25         \\
            DANN     \cite{ganin2016domain}     & 75.4          & 74.6          & 75            \\
            ADR     \cite{saito2017adversarial} & 73.3          & 74.1          & 73.7          \\
            CDAN    \cite{long2018conditional}  & 74.4          & 71.4          & 72.9          \\
            ENT  \cite{grandvalet2004semi}      & 75.4          & 75.1          & 75.25         \\
            MME   \cite{saito2019semi}          & 76.3          & 77.6          & 76.95         \\
            CLDA \cite{singh2021clda}           & \textbf{78.6} & 76.7          & 77.6          \\
            \textbf{\Method \ (Ours)}           & 78.0          & \textbf{79.0} & \textbf{78.5} \\ \hline
        \end{tabular}
\end{wraptable}

\subsubsection*{\textbf{Office-31:}}
We compare \Method's performance against other baseline approachs on the Office-31 benchmark.
We use a VGG-16 backbone for this set of experiments.
From the results in Table \ref{Tab:office31}, we observe a superior mean classification accuracy in the $3$-shot setting.

\subsubsection*{\textbf{Domain-Net:}} Table \ref{Tab:domainnet} compares the performance of \Method \ with baseline approaches, showing that \Method \ surpasses the existing baselines in both $1$-shot and $3$-shot settings.
In some scenarios like R $\rightarrow$ S and C $\rightarrow$ S, \Method \ shows a significant boost in the $1$-shot setting, outperforming even the $3$-shot accuracies of the state-of-the-art baseline approaches.
\Method's mean accuracy for the $1$-shot setting ($75.6\%$) is on par with that of the $3$-shot setting for  CDAC \cite{li2021cross} ($76.0\%$).
This shows the superiority of our approach.
Moreover, we observe that \Method \ $1$-shot is on par with $5$-shot setting of CLDA \cite{singh2021clda} ($76.7\%$).
In a $3$-shot setting, \Method \ ($76.94\%$) outperforms the $5$-shot setting of CLDA and is on par with CDAC ($76.9\%$).

\section{Ablation Study}


\subsubsection*{Momentum Parameter \momentum:}
The momentum parameter \momentum \ used in Eqn. \ref{eqn:ema} controls how the exponential moving average of the class probabilities of an unlabeled sample is updated in the mapping $\mathcal{P}$ during training.
Setting the momentum \momentum \ to $1.0$ disables the averaging function, and thus $\mathcal{P}(\operatorname{ID}(x_{i}))$ contains the latest sharpened value of sharpened soft pseudo-label $\pi(\tilde{y}_{i})$ (computed using Eqn. \ref{eqn:pseudo_label} and Eqn. \ref{eqn:sharpening}).
On the other hand, lower values of \momentum \ signify a more conservative approach, wherein weightage to the current sharpened soft pseudo-label is less.

Through our experiments, and as shown in Table \ref{Tab:ablation_momentum}, we found that setting \momentum $ =0.7$ for both Office-Home and DomainNet benchmarks worked well across different domain adaptation scenarios

\begin{wraptable}[7]{r}{7cm}
    \vspace{-2.5em}
    \centering
    \caption{Impact of threshold $\gamma$}
    \label{Tab:threshold}
    \begin{tabular}{@{}c|cc|cc@{}}
        \hline
        \multicolumn{1}{c|}{Threshold}  & \multicolumn{2}{c|}{Office-Home} & \multicolumn{2}{c}{DomainNet}                                         \\ $\gamma$
                                            & C $\rightarrow$ P                & R $\rightarrow$ C             & R $\rightarrow$ C & S $\rightarrow$ P \\ \hline

        0.9              & 82.26                            & 67.36                         & \textbf{79.23}    & \textbf{74.48}    \\

        0.8                                 & \textbf{83.82}                   & \textbf{69.33}                & 72.30                 & 70.36             \\

        0.7             & 82.70                            & 65.73                         & 68.66            & 66.90            \\ \hline
    \end{tabular}
\end{wraptable}
\subsubsection*{Injection Threshold \threshold:}
In Table \ref{Tab:threshold}, we compare the effect in the performance of our model on different domain adaptation scenarios for different values of threshold \threshold.
A higher value of threshold indicates that we will be more conservative in injecting samples into the labeled target dataset, and vice versa.
In our experiments, we observe better setting $\gamma=0.8$ for Office-Home and $\gamma=0.9$ for DomainNet gives us the best performance.



\begin{table}[t]
    \caption{(a) Impact of different values of momentum parameter $\rho$ used to compute the exponential moving average in Eqn. \ref{eqn:ema}. Results are from the Office-Home benchmark. (b) Different components of \Method. The last row uses EMA to add and remove samples from $\labeledTargetDataset$}
    \label{tab:1}
    \begin{subtable}[h]{0.32\textwidth}
        \caption{}
        \label{Tab:ablation_momentum}
        \centering
        
        \begin{tabular}{@{}c|cc@{}}
            \hline
            \multicolumn{1}{c|}{Momentum}                     & \multicolumn{2}{c}{Office-Home}                                         \\ $\rho$
                                                       & C $\rightarrow$ P                & P $\rightarrow$ R \\ \hline
            1.0             & 82.2              & 81.0              \\
            0.9             & 82.1              & 81.1              \\
            0.7             & \textbf{83.8}     & \textbf{81.9}     \\
            0.5             & 82.3              & 81.3              \\
            0.3             & 81.6              & 80.6              \\
            0.1             & 79.7              & 79.4              \\ \hline
        \end{tabular}
    \end{subtable} 
    \begin{subtable}[h]{0.68\textwidth}
        \centering
        \caption{}
        \label{Tab:losses}
        \begin{tabular}{@{}l|cc|cc@{}}
            \hline
            \multirow{2}{*}{Losses}                    & \multicolumn{2}{c|}{Office-Home} & \multicolumn{2}{c}{DomainNet}                                         \\
                                                       & C $\rightarrow$ P                & R $\rightarrow$ C             & R $\rightarrow$ C & S $\rightarrow$ P \\ \hline
            $\contrastiveLoss$                               & 82.42                            & 68.35                         & 68.15             & 67.75             \\
            $\contrastiveLoss$ + $\pawsLoss$                & 83.44                            & 69.04                         & 70.43             & 68.72             \\
            $\contrastiveLoss$ + $\consistencyLoss$                & 83.06                            & 69.02                         & 69.60             & 67.79             \\
            $\consistencyLoss$ + $\pawsLoss$                & 74.84                            & 65.64                         & 59.90             & 60.37             \\
            $\contrastiveLoss$ + $\consistencyLoss$ + $\pawsLoss$ & 82.23                  & 66.35                & 71.88    & 70.66   \\ 
            $\contrastiveLoss$ + $\consistencyLoss$ + $\pawsLoss$ EMA & \textbf{83.82}                   & \textbf{69.33}                & \textbf{79.23}    & \textbf{74.48}    \\ \hline
        \end{tabular}
    \end{subtable}
\end{table}
\subsubsection*{Effect of different components of \Method:}

Table \ref{Tab:losses} Row 1 presents the top-$1$ accuracies on different tasks using only $\contrastiveLoss$ to train the model.
We observe an increase in the top-$1$ accuracies when using $\pawsLoss$ (Row 2) and $\consistencyLoss$ (Row 3) with the supervised contrastive loss.
The performance gain when using $\pawsLoss$ and $\consistencyLoss$ separately with $\contrastiveLoss$ is not significant.
However, when used together (Row 6), we achieve the best performance with up to $9.6\%$ and $6.7\%$ gain in the top-$1$ accuracy for R $\rightarrow$ C and S $\rightarrow$ P, respectively, in the DomainNet benchmark.

Table \ref{Tab:losses} (Rows 5 and 6) shows the effect of using EMA for pseudo-label injection.
It is evident that using EMA helps improve the overall performance of the model.
Lastly, we report the results without using $\contrastiveLoss$ for domain alignment.
The stark difference in performance hints at the importance of $\contrastiveLoss$ in inter-domain alignment.
We also point out that even without the explicit domain alignment loss, \Method's performance is on par with the $1$-shot performance of D{\footnotesize E}C{\footnotesize O}T{\footnotesize A} in both DomainNet scenarios.
Therefore, it is evident that even without an explicit domain-alignment loss, SPI can leverage the similarity between the feature representations to achieve domain alignment to some extent.

\subsubsection*{Removal of Injected Samples:}
The key idea of our proposed method is the injection and removal of samples from the labeled target dataset $\labeledTargetDataset$.
In this section, we study the effect of removing injected samples from $\labeledTargetDataset$.
We posit that, over the course of training, as the model assigns the pseudo-label using Eqn. \ref{eqn:pseudo_label}, injected samples in $\labeledTargetDataset$ could be incorrectly labeled and need to be removed.
From the results presented in Table \ref{Tab:no_remove}, it is evident that removing injected samples from $\labeledTargetDataset$ provides a boost in performance, which is more profound in DomainNet scenarios.

\subsubsection*{Pseudo-label Injection Interval:}
In \Method, we choose to inject the samples into $\labeledTargetDataset$ at the end of every epoch.
We empirically show in Table \ref{Tab:iter_vs_epoch} that this setting performs better than injecting/removing samples after every iteration.

\begin{table}[t]
    \caption{Performance comparison of \Method \ to study the effect of (a) removing injected samples from labeled target dataset $\labeledTargetDataset$ and, (b) when samples are injected per epoch vs. per iteration}
    \begin{subtable}[h]{0.5\textwidth}
    \centering

    \caption{}
    \label{Tab:no_remove}
    \begin{tabular}{@{}c|cc|cc@{}}
        \hline
        \multirow{2}{*}{Removal} & \multicolumn{2}{c|}{Office-Home} & \multicolumn{2}{c}{DomainNet}                                         \\
                                & C $\rightarrow$ P                & R $\rightarrow$ C             & R $\rightarrow$ C & S $\rightarrow$ P \\ \hline



        No              & 82.99                            & 68.26                         & 70.28             & 68.82             \\

        Yes            & \textbf{83.82}                   & \textbf{69.33}                & \textbf{79.23}    & \textbf{74.48}    \\ \hline
    \end{tabular}

    \end{subtable}
    \begin{subtable}[h]{0.5\textwidth}

    \centering

    \caption{}
    \label{Tab:iter_vs_epoch}
    \begin{tabular}{@{}c|cc|cc@{}}
        \hline
        \multicolumn{1}{c|}{Updation} & \multicolumn{2}{c|}{Office-Home} & \multicolumn{2}{c}{DomainNet}                                         \\ Method 
                                & C $\rightarrow$ P                & R $\rightarrow$ C             & R $\rightarrow$ C & S $\rightarrow$ P \\ \hline

        Iter  & 83.44             & 69.04             & 75.23                 & 72.07             \\

        Epoch & \textbf{83.82}    & \textbf{69.33}    & \textbf{79.23}    & \textbf{74.48}    \\ \hline
    \end{tabular}

\end{subtable}
\end{table}

\section{Conclusions}
In this work, we presented an end-to-end framework, \Method \, which leverages feature level similarity between the unlabeled and labeled samples across domains to pseudo-label the unlabeled data.
We introduced a novel loss function that includes 1) a supervised contrastive loss for inter-domain alignment, 2) an instance-level similarity loss to pull the unlabeled samples closer to the similar labeled samples, and, 3) an intra-domain consistency loss that clusters similar unlabeled target samples.
We introduced a pseudo-labeling technique to inject (remove) confident (unconfident) samples into (from) the labeled target dataset.
We performed extensive experiments and ablations to verify the efficacy of our method.
Our framework \Method \ achieved state-of-the-art accuracies on the popular domain adaptation benchmarks.



%
%
\bibliographystyle{splncs04}
\bibliography{egbib}

\begin{thebibliography}{10}
\providecommand{\url}[1]{\texttt{#1}}
\providecommand{\urlprefix}{URL }
\providecommand{\doi}[1]{https://doi.org/#1}

\bibitem{assran2021paws}
Assran, M., Caron, M., Misra, I., Bojanowski, P., Joulin, A., Ballas, N.,
  Rabbat, M.: Semi-supervised learning of visual features by non-parametrically
  predicting view assignments with support samples. In: Proceedings of the
  IEEE/CVF International Conference on Computer Vision. pp. 8443--8452 (2021)

\bibitem{bachman2019learning}
Bachman, P., Hjelm, R.D., Buchwalter, W.: Learning representations by
  maximizing mutual information across views. Advances in neural information
  processing systems  \textbf{32} (2019)

\bibitem{berthelot2019mixmatch}
Berthelot, D., Carlini, N., Goodfellow, I., Papernot, N., Oliver, A., Raffel,
  C.A.: Mixmatch: A holistic approach to semi-supervised learning. Advances in
  Neural Information Processing Systems  \textbf{32} (2019)

\bibitem{caron2020unsupervised}
Caron, M., Misra, I., Mairal, J., Goyal, P., Bojanowski, P., Joulin, A.:
  Unsupervised learning of visual features by contrasting cluster assignments.
  Advances in Neural Information Processing Systems  \textbf{33},  9912--9924
  (2020)

\bibitem{caron2021emerging}
Caron, M., Touvron, H., Misra, I., J{\'e}gou, H., Mairal, J., Bojanowski, P.,
  Joulin, A.: Emerging properties in self-supervised vision transformers. In:
  Proceedings of the IEEE/CVF International Conference on Computer Vision. pp.
  9650--9660 (2021)

\bibitem{chen2019joint}
Chen, C., Chen, Z., Jiang, B., Jin, X.: Joint domain alignment and
  discriminative feature learning for unsupervised deep domain adaptation. In:
  Proceedings of the AAAI conference on artificial intelligence. vol.~33, pp.
  3296--3303 (2019)

\bibitem{tachet2020domain}
Tachet~des Combes, R., Zhao, H., Wang, Y.X., Gordon, G.J.: Domain adaptation
  with conditional distribution matching and generalized label shift. Advances
  in Neural Information Processing Systems  \textbf{33},  19276--19289 (2020)

\bibitem{cubuk2020randaugment}
Cubuk, E.D., Zoph, B., Shlens, J., Le, Q.V.: Randaugment: Practical automated
  data augmentation with a reduced search space. In: Proceedings of the
  IEEE/CVF Conference on Computer Vision and Pattern Recognition Workshops. pp.
  702--703 (2020)

\bibitem{deng2009imagenet}
Deng, J., Dong, W., Socher, R., Li, L.J., Li, K., Fei-Fei, L.: Imagenet: A
  large-scale hierarchical image database. In: 2009 IEEE conference on computer
  vision and pattern recognition. pp. 248--255. Ieee (2009)

\bibitem{ganin2015unsupervised}
Ganin, Y., Lempitsky, V.: Unsupervised domain adaptation by backpropagation.
  In: International conference on machine learning. pp. 1180--1189. PMLR (2015)

\bibitem{ganin2016domain}
Ganin, Y., Ustinova, E., Ajakan, H., Germain, P., Larochelle, H., Laviolette,
  F., Marchand, M., Lempitsky, V.: Domain-adversarial training of neural
  networks. The journal of machine learning research  \textbf{17}(1),
  2096--2030 (2016)

\bibitem{grandvalet2004semi}
Grandvalet, Y., Bengio, Y.: Semi-supervised learning by entropy minimization.
  Advances in neural information processing systems  \textbf{17} (2004)

\bibitem{han2020automatically}
Han, K., Rebuffi, S.A., Ehrhardt, S., Vedaldi, A., Zisserman, A.: Automatically
  discovering and learning new visual categories with ranking statistics. arXiv
  preprint arXiv:2002.05714  (2020)

\bibitem{he2020momentum}
He, K., Fan, H., Wu, Y., Xie, S., Girshick, R.: Momentum contrast for
  unsupervised visual representation learning. In: Proceedings of the IEEE/CVF
  conference on computer vision and pattern recognition. pp. 9729--9738 (2020)

\bibitem{he2016deep}
He, K., Zhang, X., Ren, S., Sun, J.: Deep residual learning for image
  recognition. In: Proceedings of the IEEE conference on computer vision and
  pattern recognition. pp. 770--778 (2016)

\bibitem{jiang2020bidirectional}
Jiang, P., Wu, A., Han, Y., Shao, Y., Qi, M., Li, B.: Bidirectional adversarial
  training for semi-supervised domain adaptation. In: IJCAI. pp. 934--940
  (2020)

\bibitem{kang2019contrastive}
Kang, G., Jiang, L., Yang, Y., Hauptmann, A.G.: Contrastive adaptation network
  for unsupervised domain adaptation. In: Proceedings of the IEEE/CVF
  Conference on Computer Vision and Pattern Recognition. pp. 4893--4902 (2019)

\bibitem{khosla2020supervised}
Khosla, P., Teterwak, P., Wang, C., Sarna, A., Tian, Y., Isola, P., Maschinot,
  A., Liu, C., Krishnan, D.: Supervised contrastive learning. Advances in
  Neural Information Processing Systems  \textbf{33},  18661--18673 (2020)

\bibitem{kim2020attract}
Kim, T., Kim, C.: Attract, perturb, and explore: Learning a feature alignment
  network for semi-supervised domain adaptation. In: European conference on
  computer vision. pp. 591--607. Springer (2020)

\bibitem{li2021learning}
Li, B., Wang, Y., Zhang, S., Li, D., Keutzer, K., Darrell, T., Zhao, H.:
  Learning invariant representations and risks for semi-supervised domain
  adaptation. In: Proceedings of the IEEE/CVF Conference on Computer Vision and
  Pattern Recognition. pp. 1104--1113 (2021)

\bibitem{li2021cross}
Li, J., Li, G., Shi, Y., Yu, Y.: Cross-domain adaptive clustering for
  semi-supervised domain adaptation. In: Proceedings of the IEEE/CVF Conference
  on Computer Vision and Pattern Recognition. pp. 2505--2514 (2021)

\bibitem{long2015learning}
Long, M., Cao, Y., Wang, J., Jordan, M.: Learning transferable features with
  deep adaptation networks. In: International conference on machine learning.
  pp. 97--105. PMLR (2015)

\bibitem{long2018conditional}
Long, M., Cao, Z., Wang, J., Jordan, M.I.: Conditional adversarial domain
  adaptation. Advances in neural information processing systems  \textbf{31}
  (2018)

\bibitem{muller2019does}
M{\"u}ller, R., Kornblith, S., Hinton, G.E.: When does label smoothing help?
  Advances in neural information processing systems  \textbf{32} (2019)

\bibitem{van2018representation}
Van~den Oord, A., Li, Y., Vinyals, O.: Representation learning with contrastive
  predictive coding. arXiv e-prints pp. arXiv--1807 (2018)

\bibitem{paul2020domain}
Paul, S., Tsai, Y.H., Schulter, S., Roy-Chowdhury, A.K., Chandraker, M.: Domain
  adaptive semantic segmentation using weak labels. In: European conference on
  computer vision. pp. 571--587. Springer (2020)

\bibitem{peng2019moment}
Peng, X., Bai, Q., Xia, X., Huang, Z., Saenko, K., Wang, B.: Moment matching
  for multi-source domain adaptation. In: Proceedings of the IEEE/CVF
  international conference on computer vision. pp. 1406--1415 (2019)

\bibitem{qin2021contradictory}
Qin, C., Wang, L., Ma, Q., Yin, Y., Wang, H., Fu, Y.: Contradictory structure
  learning for semi-supervised domain adaptation. In: Proceedings of the 2021
  SIAM International Conference on Data Mining (SDM). pp. 576--584. SIAM (2021)

\bibitem{qiu2021meta}
Qiu, S., Zhu, C., Zhou, W.: Meta self-learning for multi-source domain
  adaptation: A benchmark. In: Proceedings of the IEEE/CVF International
  Conference on Computer Vision. pp. 1592--1601 (2021)

\bibitem{saenko2010adapting}
Saenko, K., Kulis, B., Fritz, M., Darrell, T.: Adapting visual category models
  to new domains. In: European conference on computer vision. pp. 213--226.
  Springer (2010)

\bibitem{saito2019semi}
Saito, K., Kim, D., Sclaroff, S., Darrell, T., Saenko, K.: Semi-supervised
  domain adaptation via minimax entropy. In: Proceedings of the IEEE/CVF
  International Conference on Computer Vision. pp. 8050--8058 (2019)

\bibitem{saito2017adversarial}
Saito, K., Ushiku, Y., Harada, T., Saenko, K.: Adversarial dropout
  regularization. arXiv preprint arXiv:1711.01575  (2017)

\bibitem{shen2018wasserstein}
Shen, J., Qu, Y., Zhang, W., Yu, Y.: Wasserstein distance guided representation
  learning for domain adaptation. In: Thirty-second AAAI conference on
  artificial intelligence (2018)

\bibitem{shui2020beyond}
Shui, C., Chen, Q., Wen, J., Zhou, F., Gagn{\'e}, C., Wang, B.: Beyond
  h-divergence: Domain adaptation theory with jensen-shannon divergence  (2020)

\bibitem{singh2021clda}
Singh, A.: Clda: Contrastive learning for semi-supervised domain adaptation.
  Advances in Neural Information Processing Systems  \textbf{34} (2021)

\bibitem{sohn2020fixmatch}
Sohn, K., Berthelot, D., Carlini, N., Zhang, Z., Zhang, H., Raffel, C.A.,
  Cubuk, E.D., Kurakin, A., Li, C.L.: Fixmatch: Simplifying semi-supervised
  learning with consistency and confidence. Advances in Neural Information
  Processing Systems  \textbf{33},  596--608 (2020)

\bibitem{sun2016return}
Sun, B., Feng, J., Saenko, K.: Return of frustratingly easy domain adaptation.
  In: Proceedings of the AAAI Conference on Artificial Intelligence. vol.~30
  (2016)

\bibitem{tian2020contrastive}
Tian, Y., Krishnan, D., Isola, P.: Contrastive multiview coding. In: European
  conference on computer vision. pp. 776--794. Springer (2020)

\bibitem{venkateswara2017deep}
Venkateswara, H., Eusebio, J., Chakraborty, S., Panchanathan, S.: Deep hashing
  network for unsupervised domain adaptation. In: Proceedings of the IEEE
  conference on computer vision and pattern recognition. pp. 5018--5027 (2017)

\bibitem{wang2022cross}
Wang, R., Wu, Z., Weng, Z., Chen, J., Qi, G.J., Jiang, Y.G.: Cross-domain
  contrastive learning for unsupervised domain adaptation. IEEE Transactions on
  Multimedia  (2022)

\bibitem{rw2019timm}
Wightman, R.: Pytorch image models.
  \url{https://github.com/rwightman/pytorch-image-models} (2019).
  \doi{10.5281/zenodo.4414861}

\bibitem{wu2019domain}
Wu, Y., Winston, E., Kaushik, D., Lipton, Z.: Domain adaptation with
  asymmetrically-relaxed distribution alignment. In: International Conference
  on Machine Learning. pp. 6872--6881. PMLR (2019)

\bibitem{yang2021deep}
Yang, L., Wang, Y., Gao, M., Shrivastava, A., Weinberger, K.Q., Chao, W.L.,
  Lim, S.N.: Deep co-training with task decomposition for semi-supervised
  domain adaptation. In: Proceedings of the IEEE/CVF International Conference
  on Computer Vision. pp. 8906--8916 (2021)

\bibitem{yao2015semi}
Yao, T., Pan, Y., Ngo, C.W., Li, H., Mei, T.: Semi-supervised domain adaptation
  with subspace learning for visual recognition. In: Proceedings of the IEEE
  conference on Computer Vision and Pattern Recognition. pp. 2142--2150 (2015)

\bibitem{zhang2021flexmatch}
Zhang, B., Wang, Y., Hou, W., Wu, H., Wang, J., Okumura, M., Shinozaki, T.:
  Flexmatch: Boosting semi-supervised learning with curriculum pseudo labeling.
  Advances in Neural Information Processing Systems  \textbf{34} (2021)

\bibitem{zhao2019learning}
Zhao, H., Des~Combes, R.T., Zhang, K., Gordon, G.: On learning invariant
  representations for domain adaptation. In: International Conference on
  Machine Learning. pp. 7523--7532. PMLR (2019)

\end{thebibliography}
\end{document}


\pagestyle{headings}
\mainmatter
\def\ECCVSubNumber{60}  

\title{Semi-Supervised Domain Adaptation by Similarity based Pseudo-label Injection: Supplementary Material} 

\titlerunning{SPI}
%

\author{
  Abhay Rawat\inst{1,2}\orcidlink{0000-0003-3406-2149}\index{Rawat, Abhay} \and
  Isha Dua\inst{2}\orcidlink{0000-0001-5494-059X} \and
  Saurav Gupta\inst{2}\orcidlink{0000-0003-2059-1760} \and
  Rahul Tallamraju\inst{2}\orcidlink{0000-0002-8087-2225}
}
\authorrunning{A. Rawat et al.}

\institute{International Institute of Information Technology, Hyderabad, India \and
  Mercedes-Benz Research and Development India, Bengaluru, India\\
  \email{\{firstname.lastname\}@mercedes-benz.com}
}

\maketitle


\begin{algorithm}
  \DontPrintSemicolon
  \caption{Pseudocode for \Method}
  \label{alg:method}
  \KwIn{Labeled source dataset $\labeledSourceDataset$, labeled target dataset $\labeledTargetDataset$ \ and unlabeled target dataset $\unlabeledTargetDataset$. Mapping of image IDs to pseudo-labels $\mathcal{P}$}
  \Parameter{Momentum parameter $\rho$, Batch size $B_{u}$, number of classes $C$, number of support samples per class $\eta_{sup}$, number of local $\eta_{l}$ and global $\eta_{g}$ views. Number of epochs $\epsilon$, Number of iterations per epoch $N_{iters}$}

  \For{$i = 1,\dots,\epsilon$}{
    \For{$j = 1,\dots,N_{iters}$}{
      Sample support set $x_{sup}$ from $\labeledSourceDataset$ and $\labeledTargetDataset$\;
      Sample unlabeled samples $x$ from $\unlabeledTargetDataset$\;
      Compute feature representations: \\
      \nonl $\quad z_{sup} = \mathcal{F}(x_{sup})$ \\
      \nonl $\quad z = \mathcal{F}(x)$\;
      Compute soft pseudo-label: \\
      \nonl $\quad \tilde{y} = \sigma_{\tau}(\hat{z} \cdot \hat{z}_{sup}^{\top}) \ y_{sup}$ \quad where, \quad $\hat{z}_{k} ( = {z_{k}} / {\lVert z_{k} \rVert} )$\;
      Update $\mathcal{P}$:\\
      \For(){$l = 1,\dots,B_{u}$}{
        $\pi(\tilde{y_{l}}) = \frac{\tilde{y_{l}}^{1 / \tau}}{\sum_{m=1}^{C} \tilde{y}_{lm}^{1 / \tau}}$\;
        $\mathcal{P}(\operatorname{ID}(x_{l})) \leftarrow \rho \ \pi(\tilde{y}_{l}) + (1 - \rho) \ \mathcal{P}(\operatorname{ID}(x_{l}))$\;
      }
      Compute $\mathcal{L}_{\Method} = \lambda_{1} \contrastiveLoss + \lambda_{2} \pawsLoss + \lambda_{3} \consistencyLoss + \lambda_{4} \classifierLoss$\;
      Compute gradients $\nabla \Loss{\Method}$ for weight update.\;
      \nonl Update feature extractor and classifier weights, $\Theta_{\mathcal{F}}$ and $\Theta_{\mathcal{H}}$\;
    }
    \tcc{Find samples to inject into $\labeledTargetDataset$}
    $I = \{ (x, \argmax \mathcal{P}(\operatorname{ID}(x)) \ \vert \ x \in  T \land \max{\mathcal{P}(\operatorname{ID}(x))} \geq \gamma \}$\;
    \tcc{Find samples to remove from $\labeledTargetDataset$}
    $R = \{ (x, y) \ \vert \ x \in  (\hat{T} \setminus \hat{T}_{0}) \land \max{\mathcal{P}(\operatorname{ID}(x))} < \gamma \}$\;
    \tcc{Update $\labeledTargetDataset$}
    $
      \hat{T} =
      \begin{cases}
        (\hat{T} \setminus R ) \cup I & \text{if} \ i \geq W \\
        \hat{T}                              & \text{otherwise}
      \end{cases}
    $
  }
\end{algorithm}

\section{Implementation Details}
We present the pseudo-code of our approach in Algorithm \ref{alg:method}.
For training our model, we set the values of the temperature parameter $\tau$ in Eqn. 2 and 4 (from the the main paper) as $0.1$ and $0.3$ respectively.
The temperature parameter used in Eqn. 3 set to $0.7$ at the beginning of training and is annealed using a cosine scheduler to a minimum value of $0.25$ throught the training process.

\section{Feature Space Analysis}
\begin{figure}[h]
  \centering
  \includegraphics[width=0.7\columnwidth, trim=0 0 0 0, clip=true]{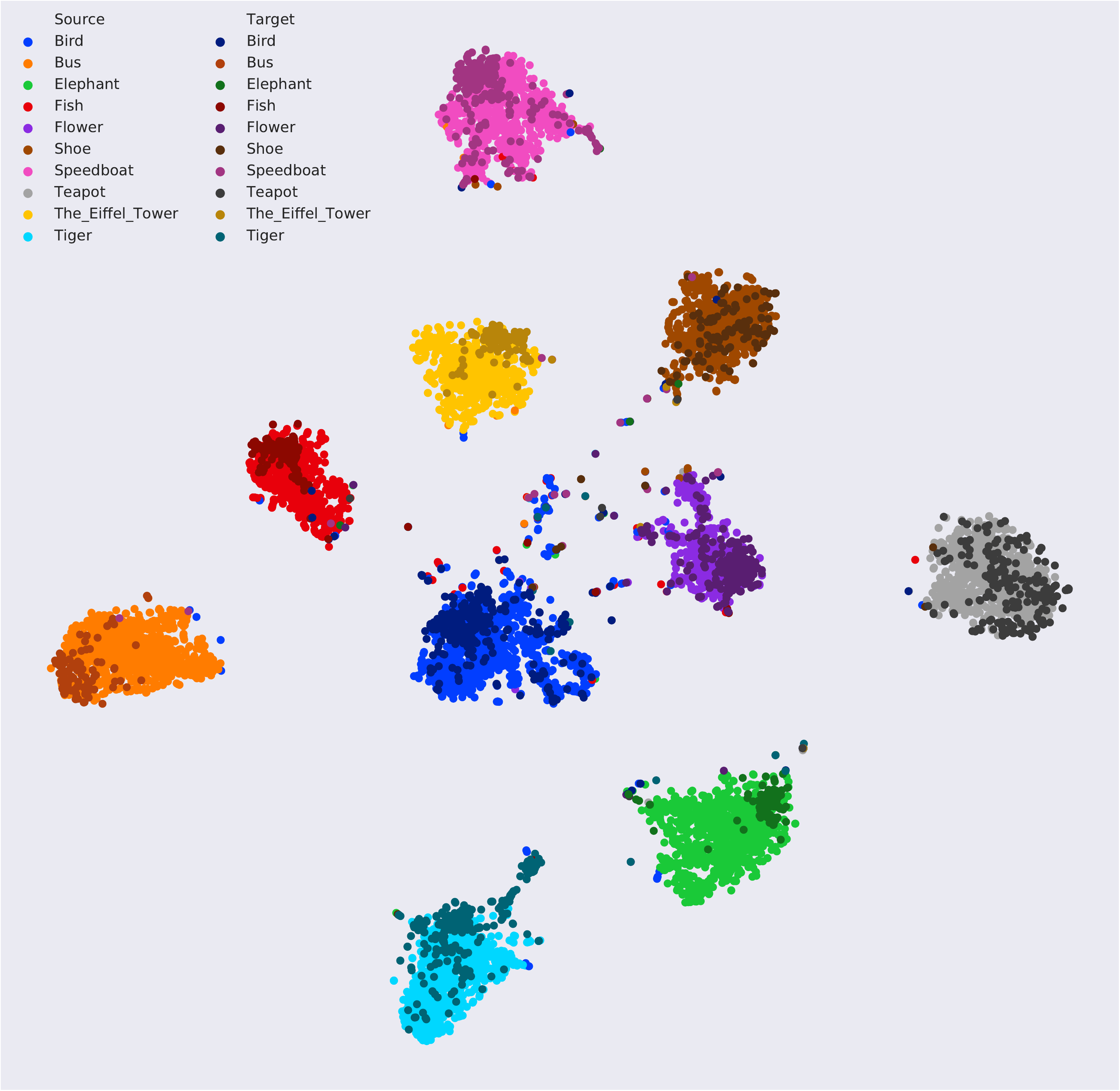}
  \caption{t-SNE visualization of features. We show the features for $10$ representative classes from the R $\rightarrow$ C scenario from the DomainNet benchmark.}
  \label{fig:tsne}
\end{figure}

\begin{figure}[t]
  \centering
  \includegraphics[width=\textwidth, trim=0 4cm 0 0, clip=true]{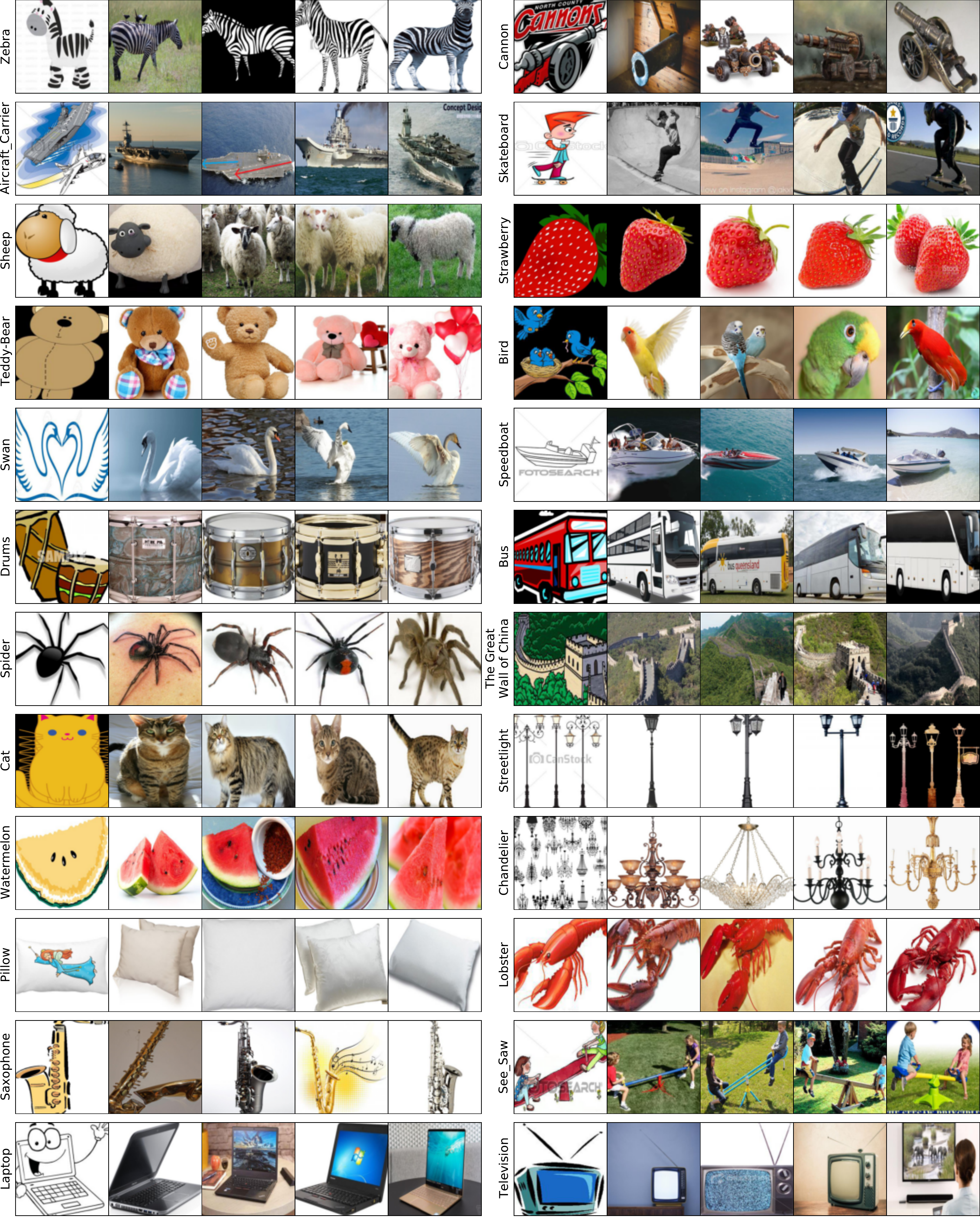}
  \caption{Nearest neighbours from the source domain (Real), of the target domain (Clipart) sample. The $1$st column is the target domain sample used for query.}
  \label{fig:ret}
\end{figure}

In Figure \ref{fig:ret}, we show the top-$4$ nearest neighbours in the source domain (Real), of the images from the target domain (Clipart).
In Figure \ref{fig:tsne}, we show the t-SNE plot of the feature distribution in the R $\rightarrow$ C scenario from the DomainNet benchmark.
We choose $10$ representative classes for visualization as done in \cite{li2021cross}.

\begin{table*}[t]
    \centering
    \caption{Accuracy on DomainNet under the 5-shot setting using ResNet-34 backbone.}
    \label{tab:five_shot}
    \begin{tabular}{@{}c|c|ccccccc|c@{}}
    \hline
    Net                       & Method  & R→C  & R→P  & P→C           & C→S           & S→P           & R→S           & P→R           & MEAN \\ \hline
    \multirow{8}{*}{Resnet34} & S+T     & 64.5 & 63.1 & 64.2          & 59.2          & 60.4          & 56.2          & 75.7          & 63.3 \\
                              & DANN \cite{ganin2016domain}    & 63.7 & 62.9 & 60.5          & 55            & 59.5          & 55.8          & 72.6          & 61.4 \\
                              & ENT  \cite{grandvalet2004semi}     & 77.1 & 71   & 75.7          & 61.9          & 66.2          & 64.6          & 81.1          & 71.1 \\
                              & MME   \cite{saito2019semi}    & 75.5 & 70.4 & 74            & 65            & 68.2          & 65.5          & 79.9          & 71.2 \\
                              & APE  \cite{kim2020attract}     & 77.7 & 73   & 76.9          & 67            & 71.4          & 68.8          & 80.5          & 73.6 \\
                              & CDAC   \cite{li2021cross}    & \textbf{80.8} & 75.3 & 79.9          & 72.1          & 74.7          & 72.9          & 83.2          & 76.9 \\
                              & CLDA  \cite{singh2021clda}          & 80.3 & 76.0 & 77.8          & 71.6          & 74.5          & 72.9          & \textbf{84.0} & 76.7 \\
                              & \Method & 79.7 & \textbf{76.2}    & \textbf{80.1} & \textbf{74.1} & \textbf{75.4} & \textbf{74.0} & 83.6          & \textbf{77.6}    \\ \hline
    \end{tabular}
    \end{table*}
\begin{table*}[t]
    \centering
    \caption{Accuracy on DomainNet under the 10-shot setting using ResNet-34 backbone.}
    \label{tab:ten_shot}
    \begin{tabular}{c|c|ccccccc|c}
        \hline
    Net                       & Method  & R→C           & R→P  & P→C           & C→S           & S→P           & R→S  & P→R           & MEAN \\
    \hline
    \multirow{9}{*}{Resnet34} & S+T     & 68.5          & 66.4 & 69.2          & 64.8          & 64.2          & 60.7 & 77.3          & 67.3 \\
                              & DANN     \cite{ganin2016domain} & 70            & 64.5 & 64            & 56.9          & 60.7          & 60.5 & 75.9          & 64.6 \\
                              & ENT   \cite{grandvalet2004semi}     & 79            & 72.9 & 78            & 68.9          & 68.4          & 68.1 & 82.6          & 74   \\
                              & MME     \cite{saito2019semi}    & 77.1          & 71.9 & 76.3          & 67            & 69.7          & 67.8 & 81.2          & 73   \\
                              & APE    \cite{kim2020attract}    & 79.8          & 75.1 & 78.9          & 70.5          & 73.6          & 70.8 & 82.9          & 75.9 \\
                              & CDAC   \cite{li2021cross}   & 83.1          & 77.2 & 81.7          & 74.3          & 76.3          & 74.6 & 84.7          & 78.9 \\
                              & D{\footnotesize E}C{\footnotesize O}T{\footnotesize A}  \cite{yang2021deep}  & \textbf{81.8} & 75.1 & 81.3          & 73.7          & 73.4          & 73.7 & 80.7          & 77.1 \\
                              & CLDA   \cite{singh2021clda}         & 81.2          & \textbf{77.7} & 80.3          & 74.1          & \textbf{77.1} & 74.1 & \textbf{85.1}          & 78.5 \\
                              & \Method & 81.7          & 77.2    & \textbf{82.1} & \textbf{76.2} & 76.2          & \textbf{75.2}    & 85.0 &     \textbf{79.1} \\
                              \hline
    \end{tabular}
    \end{table*}



\section{Results with more shots}
We conducted experiments to analyze the performance of \Method \ when provided with more labeled samples in the target domain.
In Table \ref{tab:five_shot} and \ref{tab:ten_shot}, we compare the performance of our method with the existing approaches on scenarios in the DomainNet benchmark.

\section{Different training splits for the labeled target data}

\begin{wraptable}[6]{r}{5cm}
\vspace{-3em}
\caption{Performance of \Method \ with different training splits}
\label{tab:splits}
\begin{tabular}{@{}l|llll@{}}
\hline
Splits                 & S0                  & S1                  & S2                  & S3                  \\ \hline
 Accuracy &  69.33 &  71.82 &  69.84 &  70.19 \\ \hline
\end{tabular}
\end{wraptable}

We compare the performance of our method - \Method \ with different training splits for labeled target data. This is to show that our approach is fairly invariant to the labeled target samples. In Table \ref{tab:splits}, we observe that our method is invariant to randomly chosen three training splits (S0, S1, S2, S3) for the labeled target data. S0 denotes the split used in our main paper (The same split as used in MME \cite{saito2019semi}). S1-S3 were generated randomly.

\begin{figure}
  \centering
  \includegraphics[width=0.6\textwidth, trim=0 0 0 0, clip=true]{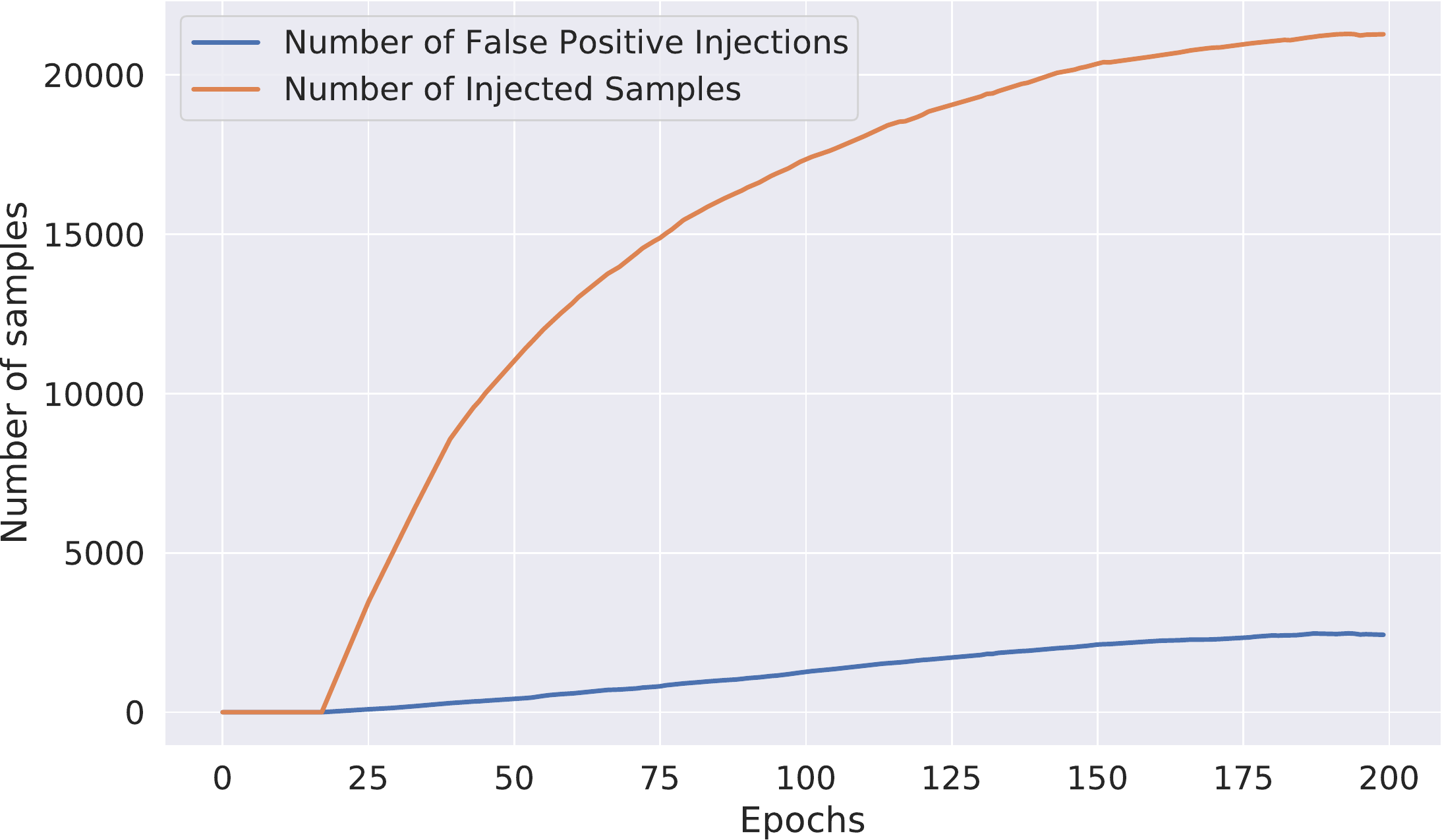}
  \caption{Number of injected samples over the course of training in R $\rightarrow$ P scenario of DomainNet.}
  \label{fig:injections}
\end{figure}

\section{Analysis for Injected Samples}

In Figure \ref{fig:injections}, we show the number of injected samples over the course of training.
We also show the number of false positive samples that have been injected into the labeled target dataset $\labeledTargetDataset$.

\clearpage
%
%
\bibliographystyle{splncs04}
\bibliography{egbib}